\documentclass[onecolumn,draftcls]{IEEEtran}

\usepackage{booktabs}

\usepackage{natbib}
\usepackage{bbm}
\usepackage[mathscr]{eucal}
\usepackage[cmex10]{amsmath}
\usepackage{epsfig,epsf,psfrag}
\usepackage{amssymb,amsmath,amsthm,amsfonts,latexsym,bm}
\usepackage{amsmath,graphicx,bm,xcolor,url}
\usepackage[caption=false]{subfig} 
\usepackage{fixltx2e}
\usepackage{array}
\usepackage{verbatim}
\usepackage{bm}
\usepackage{algorithmic, cite}
\usepackage{algorithm}
\usepackage{verbatim}
\usepackage{textcomp}
\usepackage{mathrsfs}
\usepackage{multirow}
\usepackage{epstopdf}

\catcode`~=11 \def\UrlSpecials{\do\~{\kern -.15em\lower .7ex\hbox{~}\kern .04em}} \catcode`~=13 

\allowdisplaybreaks[1]
 
\newcommand{\nn}{\nonumber}

\newcommand{\calF}{\mathcal{F}}
\newcommand{\calG}{\mathcal{G}}
\newcommand{\calH}{\mathcal{H}}

\newcommand{\calN}{\mathcal{N}}

\newcommand{\calR}{\mathcal{R}}
\newcommand{\calS}{\mathcal{S}}

\newcommand{\calV}{\mathcal{V}}
\newcommand{\calW}{\mathcal{W}}
\newcommand{\calX}{\mathcal{X}}
\newcommand{\calY}{\mathcal{Y}}

\newcommand{\bA}{\mathbf{A}}

\newcommand{\bx}{\mathbf{x}}

\newcommand{\bbE}{\mathbb{E}}

\newcommand{\bbR}{\mathbb{R}}

\newcommand{\bbZ}{\mathbb{Z}}

\DeclareMathAlphabet{\mathbsf}{OT1}{cmss}{bx}{n}
\DeclareMathAlphabet{\mathssf}{OT1}{cmss}{m}{sl}

\DeclareSymbolFont{bsfletters}{OT1}{cmss}{bx}{n}  
\DeclareSymbolFont{ssfletters}{OT1}{cmss}{m}{n}
\DeclareMathSymbol{\bsfGamma}{0}{bsfletters}{'000}
\DeclareMathSymbol{\ssfGamma}{0}{ssfletters}{'000}
\DeclareMathSymbol{\bsfDelta}{0}{bsfletters}{'001}
\DeclareMathSymbol{\ssfDelta}{0}{ssfletters}{'001}
\DeclareMathSymbol{\bsfTheta}{0}{bsfletters}{'002}
\DeclareMathSymbol{\ssfTheta}{0}{ssfletters}{'002}
\DeclareMathSymbol{\bsfLambda}{0}{bsfletters}{'003}
\DeclareMathSymbol{\ssfLambda}{0}{ssfletters}{'003}
\DeclareMathSymbol{\bsfXi}{0}{bsfletters}{'004}
\DeclareMathSymbol{\ssfXi}{0}{ssfletters}{'004}
\DeclareMathSymbol{\bsfPi}{0}{bsfletters}{'005}
\DeclareMathSymbol{\ssfPi}{0}{ssfletters}{'005}
\DeclareMathSymbol{\bsfSigma}{0}{bsfletters}{'006}
\DeclareMathSymbol{\ssfSigma}{0}{ssfletters}{'006}
\DeclareMathSymbol{\bsfUpsilon}{0}{bsfletters}{'007}
\DeclareMathSymbol{\ssfUpsilon}{0}{ssfletters}{'007}
\DeclareMathSymbol{\bsfPhi}{0}{bsfletters}{'010}
\DeclareMathSymbol{\ssfPhi}{0}{ssfletters}{'010}
\DeclareMathSymbol{\bsfPsi}{0}{bsfletters}{'011}
\DeclareMathSymbol{\ssfPsi}{0}{ssfletters}{'011}
\DeclareMathSymbol{\bsfOmega}{0}{bsfletters}{'012}
\DeclareMathSymbol{\ssfOmega}{0}{ssfletters}{'012}

\newcommand{\eps}{\varepsilon}

\newtheorem{theorem}{Theorem} 
\newtheorem{lemma}[theorem]{Lemma}

\newtheorem{corollary}[theorem]{Corollary}

\newtheorem{remark}[theorem]{Remark}

\newcommand{\qednew}{\nobreak \ifvmode \relax \else
      \ifdim\lastskip<1.5em \hskip-\lastskip
      \hskip1.5em plus0em minus0.5em \fi \nobreak
      \vrule height0.75em width0.5em depth0.25em\fi}

\usepackage{xspace}
\usepackage[ colorlinks = true,
             linkcolor = blue,
             urlcolor  = blue,
             citecolor = red,
             anchorcolor = green,
]{hyperref}

\usepackage{cite}
\allowdisplaybreaks[1]
\usepackage{amssymb}
\usepackage{pifont}

\begin{document}
\title{On Generalization Error Bounds for Transformers}
\author{Lan V.\ Truong $\,$
         
\thanks{The author is with the Faculty of Computer Science and Engineering, Ho Chi Minh City University of Technology (HCMUT), Vietnam and with Vietnam National University Ho Chi Minh City (VNU-HCM), Vietnam. Email: \url{lantv@hcmut.edu.vn.}
}}

\maketitle

\begin{abstract}
In this paper, we establish a collection of covering number bounds for linear function classes under various norm constraints on the inputs and matrices. We then combine these results with existing covering number bounds to derive improved estimates and, based on these estimates, develop generalization error bounds for single-layer Transformers. The resulting generalization bounds improve upon several existing results in the literature and, in particular, are independent of the input sequence length. Moreover, our generalization error bound decays at the rate $O(1/\sqrt{n})$, where $n$ denotes the sample size, thereby improving upon existing bounds that scale as $O((\log n)/\sqrt{n})$. Furthermore, our covering number analysis explicitly incorporates rank constraints on the underlying matrix classes, allowing us to characterize how low-rank structures affect the metric entropy and, consequently, the resulting generalization bounds for Transformer architectures.
\end{abstract}
\section{Introduction}

Transformers are a type of neural network architecture that has revolutionized natural language processing (NLP) and other fields such as computer vision and audio processing. Introduced in the 2017 paper ``Attention is All You Need" by Vaswani et al. \citep{Vaswani2017AttentionIA}, transformers rely primarily on a mechanism called self-attention, which allows the model to weigh the importance of different words (or tokens) in a sequence, regardless of their position. The transformer has emerged as one of the most influential architectures of its era, demonstrating state-of-the-art predictive capabilities across various fields \citep{Dosovitskiy2020AnII, Wu2022MultistepSW, Vaswani2017AttentionIA, Pettersson2023ComparisonOL}. However, our understanding of transformers, particularly regarding generalization errors, remains limited. Overall, a deeper understanding of generalization error in transformers can lead to better models, more robust applications, and advancements in the field of machine learning.

Traditional deep neural networks have benefited from complexity-based generalization bounds established by classical learning theory, providing theoretical guarantees for deep learning. Researchers such as Goldberg \citep{Goldberg1993}, Bartlett \citep{Bartlett1996}, \citep{Bartlett1998A} proposed upper bounds based on VC dimension for deep neural networks (DNNs). Neyshabur et al.  \citep{Neyshabur2015} utilized Rademacher complexity to show a bound with exponential dependence on the depth for ReLU networks. In further studies, Neyshabur et al. \citep{Neyshabur2018APA} and Bartlett et al. \citep{Bartlett2017} applied PAC-Bayesian analysis and covering numbers, respectively, to derive bounds that exhibit polynomial dependence on depth. Golowich et al.  \citep{Golowich2018} presented bounds indicating (sub-linear) square-root dependence on depth for DNNs with positive-homogeneous activations like ReLU. More recently, Truong \citep{Truong2022OnRC} improved upon the bounds from Golowich et al. by demonstrating that Rademacher complexity is explicitly independent of network length for convolutional neural networks (CNNs) with certain activation functions, including ReLU, Leaky ReLU, Parametric ReLU, Sigmoid, and Tanh. This insight leads to non-vacuous generalization bounds for CNNs when classifying a limited number of image classes.

For Transformers, there has been significant progress in establishing generalization error bounds. Edelman et al. \citep{Edelman2022} established a norm-based generalization bound that grows logarithmically with the sequence length, suggesting that Transformers exhibit an inductive bias toward representing sparse functions of their inputs. Trauger and Tewari \citep{Trauger2023LI} subsequently eliminated the dependence on sequence length entirely. These results rely on covering number bounds for linear function classes \citep{Zhang2002NB} and covering bounds for compositions of functions. In this work, we develop new covering number bounds for linear function classes and combine them with existing covering bounds to obtain improved estimates over the corresponding bounds of Edelman et al. \citep{Edelman2022} and Trauger and Tewari \citep{Trauger2023LI} under the same norm constraints. In addition, our covering bounds explicitly account for rank constraints on the query, key, and value matrices, and therefore also apply when these weight matrices are not necessarily of full rank. By incorporating these refined covering bounds into the analysis of single-layer Transformers, we obtain improved generalization error bounds while allowing the analysis to capture the potential benefits of low-rank structures in Transformer architectures.

More specifically, our contributions are as follows:
\begin{itemize}
    \item We establish new covering number bounds for linear function classes of the form
\[
f(x)=Wx,
\]
where $W\in\mathcal{W}$ is subject to norm constraints together with either column-space subspace constraints or rank constraints. When the prescribed column-space subspace is $\mathbb{R}^k$, the subspace constraint becomes vacuous; similarly, when the rank constraint is set to $r_w=k$, it becomes vacuous. In these respective settings, by combining our results with existing covering number bounds, we obtain improved covering number estimates over those of Edelman et al. \citep{Edelman2022} and Trauger and Tewari \citep{Trauger2023LI} under the same norm constraints. More generally, our results provide rank-dependent covering bounds that quantify the effect of non-full-rank weight matrices. In particular, these bounds become sharper when the weight matrices have rank strictly smaller than their maximal possible rank.

    \item We apply these covering number bounds to derive sharper generalization error bounds for single-layer Transformers. In particular, the resulting bounds incorporate structural information about the weight matrices, including their rank, and are independent of the input sequence length. 

    \item We show that the resulting generalization error bounds decay at the rate $O(1/\sqrt{n})$, improving upon previous bounds of order $O((\log n)/\sqrt{n})$. The improvement follows from the sharper covering number estimates developed in the first contribution. 
\end{itemize}

Since an earlier version of this work appeared on arXiv, several recent studies have investigated low-rank structures in Transformers from
different theoretical perspectives. Kajitsuka and Sato~\citep{Kajitsuka2024} studied the expressive power of Transformers with low-rank weight matrices
and showed that even a single self-attention layer with low-rank weights can have substantial memorization and universal approximation capabilities.
Li et al.~\citep{Li2024} theoretically analyzed the emergence of low-rank and sparse structures during the training of one-layer Transformers,
showing that the gradient updates can exhibit low-rank structures under their data model. Low-rank adaptation has also received increasing
theoretical attention; for example, Jang et al.~\citep{Jang2024} analyzed LoRA training in the neural tangent kernel regime and established
conditions under which low-rank adaptation avoids spurious local minima and achieves good generalization. More recently, Pinto et al.~\citep{Pinto2025}
studied generalization bounds for neural networks with low-rank layers and showed how low-rank constraints can improve generalization guarantees
by controlling the accumulation of rank and dimensionality factors.

Our work is complementary to these developments. While the above studies focus primarily on expressivity, training dynamics, low-rank adaptation, or generalization bounds for neural networks, we focus on the metric entropy and statistical complexity of Transformer components. In particular, we derive explicit rank-dependent covering number bounds under both fixed-subspace constraints and rank-only constraints. These covering bounds yield corresponding generalization guarantees and can be combined with existing covering estimates to obtain tighter bounds under the same norm constraints. Thus, our work provides a distinct statistical learning-theoretic perspective on the role of low-rank structures in Transformer generalization, complementing existing studies on the expressivity, optimization, and adaptation properties of low-rank neural architectures.

\section{Background}
\subsection{Matrix norms} Vectors and matrices are in boldface. 
For any vector $\bx=(x_1,x_2,\cdots,x_n) \in \bbR^n$ where $\bbR$ is the field of real numbers, its induced-$L^p$ norm is defined as
\[
\|\bx\|_p=\bigg(\sum_{k=1}^n |x_k|^p\bigg)^{1/p}. 
\]
The $j$-th component of the vector $\bx$ is denoted as $x_j$ for all $j \in [n]$. 

For $\bA \in \bbR^{m\times n}$ where
\[
 \bA=\begin{bmatrix}a_{11},&a_{12},&\cdots,&a_{1n}\\ a_{21},&a_{22},&\cdots,&a_{2n}\\ \vdots& \vdots&\ddots& \vdots \\ a_{m1},& a_{m2},&\cdots, &a_{m n}\end{bmatrix}
\]
 we defined the induced-norm of matrix $\bA$ as
\[
 \|\bA\|_{p\to q}=\sup_{\bx \neq \b0} \frac{\|\bA \bx\|_q}{\|\bx\|_p}.
\]
 
The ``entry-wise" matrix norm of $A$ is defined as follows:
\[
\|A\|_{p,q}:= \bigg(\sum_{j=1}^n\bigg(\sum_{i=1}^m |a_{ij}|^p\bigg)^{\frac{1}{p}}\bigg)^{\frac{1}{q}}.
\]
For any $m \in \bbZ_+$, we use the notation $[m] := \{1,2,\ldots, m\}$.

\subsection{Generalization and Rademacher Complexity}

Let $\mathcal{H}$ be a hypothesis class with loss $\ell$ and data distribution
$\mathcal{D}$ on $\mathcal{X}\times\mathcal{Y}$.
Given a sample $\mathcal{S}=\{(x_i,y_i)\}_{i=1}^n$, the generalization gap of
$h\in\mathcal{H}$ is
\begin{align}
\Big|
\mathbb{E}_{(x,y)\sim\mathcal{D}}[\ell(h(x),y)]
-
\frac{1}{n}\sum_{i=1}^n \ell(h(x_i),y_i)
\Big|.
\end{align}

Generalization can be controlled via \emph{Rademacher complexity}. For inputs
$\{x_i\}_{i=1}^n$, the Rademacher complexity is defined as
\begin{align}
\mathcal{R}_n(\mathcal{H},\{x_i\}_{i=1}^n)
=
\frac{1}{n}
\mathbb{E}_{\boldsymbol{\sigma}}
\Big[
\sup_{h\in\mathcal{H}}
\sum_{i=1}^n \sigma_i h(x_i)
\Big],
\end{align}
where $\sigma_1,\ldots,\sigma_n$ are i.i.d.\ Rademacher random variables.

If $|\ell|\le c$, then with probability at least $1-\delta$,
\begin{align}
&\sup_{h\in\mathcal{H}}
\Big|
\mathbb{E}_{(x,y) \sim \mathcal{D}}[\ell(h(x),y)]
-
\frac{1}{n}\sum_{i=1}^n \ell(h(x_i),y_i)
\Big| \nn\\
&\qquad \le
2\mathcal{R}_n(\ell\circ\mathcal{H},\{x_i\}_{i=1}^n)
+
4c\sqrt{\tfrac{2\log(4/\delta)}{n}},
\end{align}
where $\ell \circ \calH:=\{(x,y) \mapsto l(h(x),y)| h \in \calH, (x,y) \in \calX \times \calY$. 

\subsection{Covering Numbers}

Let $\mathcal{F}=\{f:\mathbb{R}^d\to\mathbb{R}^k\}$ and $q>0$.
For a sample $\{x_i\}_{i=1}^n$, define
$N_{\infty}(\mathcal{F},\varepsilon,\{x_i\},\|\cdot\|_q)$
as the smallest $m \in \bbZ_+$ such that there exist functions
$f_1,\ldots,f_m\in\mathbb{R}^k$ satisfying
\[
\max_{i\in[n]}\min_{j\in[m]}
\|f(x_i)-f_j(x_i)\|_q
\le
\varepsilon
\quad
\forall f\in\mathcal{F}.
\]

The worst-case covering number over all samples of size $n$ is
\[
N_{\infty}(\mathcal{F},\varepsilon,n,\|\cdot\|_q)
=
\sup_{\{x_i\}_{i=1}^n}
N_{\infty}(\mathcal{F},\varepsilon,\{x_i\},\|\cdot\|_q).
\]
By \citep{ShalevShwartz2014UnderstandingML} if $\|f(x)\|\le c_x$ uniformly, we have:
\begin{align}
&\calR_n(\calF, \{x_i\}_{i=1}^n )\leq 2 \eps_{m+1}+ \frac{12}{\sqrt{n}} \sum_{j=1}^m (\eps_j-\eps_{j+1}) \nn\\
&\qquad \qquad \times \sqrt{\log N_{\infty}(\calF,\eps_j,n,\|\cdot\|_2)},
\end{align}
where $\eps_j:= c_x/2^j$ for all $j \in [m+1]$. 

\subsection{Self-Attention and Transformers}
In this work, we adopt the definitions and notation for self-attention and Transformer architectures as outlined by \citep{Edelman2022, Trauger2023LI}, and we use them consistently throughout.

Let $X\in\mathbb{R}^{T\times d}$ denote an input sequence.
A single attention head is parameterized by
$W_v\in\mathbb{R}^{d\times k}$,
$W_c\in\mathbb{R}^{k\times d}$,
and query--key matrices $W_Q,W_K \in \bbR^{d \times T}$.
Let $\sigma$ be an element-wise $L_\sigma$-Lipschitz activation with
$\sigma(0)=0$, and denote row-wise softmax by \texttt{RowSoftmax}.

The output of a single attention head is
\[
\sigma\!\left(
\text{RowSoftmax}(XW_Q W_K^\top X^\top)\,XW_v
\right)W_c.
\]

Since queries and keys appear only through their product, we equivalently
introduce a combined matrix
$W_{QK}=W_Q W_K^T \in\mathbb{R}^{d\times d}$,
yielding a parameterization independent of the sequence length $T$.

For scalar prediction tasks, a special token (e.g., \texttt{[CLS]}) is prepended.
Let $Y_{\texttt{[CLS]}}\in\mathbb{R}^d$ denote its final-layer representation.
The model output is given by
\[
w^\top Y_{\texttt{[CLS]}},
\qquad
w\in\mathbb{R}^d.
\]
\section{Rank-Dependent Covering Number Bounds for Linear Function Classes} \label{sec:theory}

In Section~3 of \citep{Zhang2002NB}, log-covering number bounds were derived for linear function classes of the form $x \mapsto w^\top x$, where $w$ is a vector. By contrast, we study linear function classes of the form $x \mapsto W^\top x$, where $W$ is a matrix, and develop corresponding log-covering number bounds. Exploiting matrix structure to obtain improved bounds is a central focus of the present section.
\subsection{Covering Numbers under Fixed Subspace Constraints}
In this subsection, we consider the case where the column space of $W$ is contained in a fixed subspace. Under this constraint, the rank of $W$ is bounded by the dimension of the subspace. We first establish the following result.
\begin{theorem}
\label{thm:main0}
Let
\[
V_W\subseteq\mathbb{R}^k
\]
be a fixed linear subspace with
\[
\dim(V_W)=r_w.
\]
Define
\[
\mathcal{W}
=
\left\{
W\in\mathbb{R}^{k\times d}:
\operatorname{col}(W)\subseteq V_W,
\ \|W\|_{2\to2}\le B_w
\right\},
\]
and let
\[
\mathcal{F}^{(2 \to 2)}
=
\left\{
f_W:x\mapsto Wx:
W\in\mathcal{W}
\right\}.
\]
Suppose that the inputs satisfy
\[
\|x_i\|_2\le B_x,
\qquad i=1,\ldots,n.
\]
Then, for every $\epsilon>0$,
\[
\log
\mathcal{N}_\infty
\left(
\mathcal{F}^{(2 \to 2)},\epsilon,n,\|\cdot\|_2
\right)
\le
r_w d
\log
\left(
1+\frac{2B_wB_x}{\epsilon}
\right).
\]
\end{theorem}
\begin{remark}
By setting $\mathcal{V}_W=\mathbb{R}^k$, we obtain a novel covering number bound for the linear class
\[
\left\{x\mapsto Wx:\|W\|_{2\to 2}\leq B_w\right\}.
\]
By combining our bound with the existing covering number bounds of Edelman et al. \citep{Edelman2022} and Trauger and Tewari \citep{Trauger2023LI}, we further obtain improved covering number estimates under the same norm constraint (cf. Lemma \ref{lem:aux2}).
\end{remark}
\begin{proof}
Let
\[
U\in\mathbb{R}^{k\times r_w}
\]
have orthonormal columns spanning $V_W$. Thus,
\[
U^\top U=I_{r_w},
\qquad
\operatorname{col}(U)=V_W.
\]

For every
\[
W\in\mathcal{W},
\]
the condition
\[
\operatorname{col}(W)\subseteq V_W
\]
implies that there exists a matrix
\[
A\in\mathbb{R}^{r_w\times d}
\]
such that
\[
W=UA.
\]
Moreover, this representation is unique, since
\[
A=U^\top W.
\]

Because $U$ has orthonormal columns,
\[
\|UA\|_{2\to2}=\|A\|_{2\to2}.
\]
Indeed,
\[
\begin{aligned}
\|UA\|_{2\to2}
&=
\sup_{\|x\|_2=1}\|UAx\|_2\\
&=
\sup_{\|x\|_2=1}\|Ax\|_2\\
&=
\|A\|_{2\to2}.
\end{aligned}
\]
Consequently,
\[
\mathcal{W}^{(2\to2)}
=
\left\{
UA:
A\in\mathbb{R}^{r_w\times d},
\ \|A\|_{2\to2}\le B_w
\right\}.
\]

Define
\[
\mathcal A
=
\left\{
A\in\mathbb{R}^{r_w\times d}:
\|A\|_{2\to2}\le B_w
\right\}.
\]

We use the standard covering-number bound for the
operator-norm ball in $\mathbb{R}^{r_w\times d}$ (e.g., Vershynin~\citep{Vershynin2018}):
for every $\delta>0$,
\begin{align}
\mathcal{N}
\left(
\mathcal A,\delta,\|\cdot\|_{2\to2}
\right)
\le
\left(
1+\frac{2B_w}{\delta}
\right)^{r_wd} \label{eq1b}.
\end{align}
Hence, there exists a finite set
\[
\mathcal A_\delta\subseteq\mathcal A
\]
such that
\[
|\mathcal A_\delta|
\le
\left(
1+\frac{2B_w}{\delta}
\right)^{r_wd}
\]
and, for every $A\in\mathcal A$, there exists
$\widetilde A\in\mathcal A_\delta$ satisfying
\[
\|A-\widetilde A\|_{2\to2}\le\delta.
\]

For every $\widetilde A\in\mathcal A_\delta$, define
\[
\widetilde W=U\widetilde A
\]
and
\[
\widetilde f(x)=\widetilde W x.
\]
Let
\[
\widetilde{\mathcal F}
=
\left\{
\widetilde f:
\widetilde A\in\mathcal A_\delta
\right\}.
\]

Consider any
\[
f_W\in\mathcal F.
\]
Write
\[
W=UA
\]
for some $A\in\mathcal A$, and choose
$\widetilde A\in\mathcal A_\delta$ such that
\[
\|A-\widetilde A\|_{2\to2}\le\delta.
\]
Then, for every $i=1,\ldots,n$,
\[
\begin{aligned}
\|f_W(x_i)-\widetilde f(x_i)\|_2
&=
\|(W-\widetilde W)x_i\|_2\\
&=
\|U(A-\widetilde A)x_i\|_2\\
&=
\|(A-\widetilde A)x_i\|_2\\
&\le
\|A-\widetilde A\|_{2\to2}
\|x_i\|_2\\
&\le
\delta B_x.
\end{aligned}
\]
Therefore,
\[
\max_{1\le i\le n}
\|f_W(x_i)-\widetilde f(x_i)\|_2
\le
\delta B_x.
\]

Choosing
\[
\delta=\frac{\epsilon}{B_x},
\]
we obtain
\[
\max_{1\le i\le n}
\|f_W(x_i)-\widetilde f(x_i)\|_2
\le\epsilon.
\]
Thus,
\begin{align}
\mathcal N_\infty
\left(
\mathcal F^{(2 \to 2)},\epsilon,n,\|\cdot\|_2
\right)
\le
\mathcal N
\left(
\mathcal A,
\frac{\epsilon}{B_x},
\|\cdot\|_{2\to2}
\right) \label{eq1c}.
\end{align}
From \eqref{eq1b} and \eqref{eq1c}, we obtain
\[
\mathcal N_\infty
\left(
\mathcal F^{(2 \to 2)},\epsilon,n,\|\cdot\|_2
\right)
\le
\left(
1+\frac{2B_wB_x}{\epsilon}
\right)^{r_wd}.
\]
Taking logarithms yields
\[
\log
\mathcal N_\infty
\left(
\mathcal F^{(2 \to 2)},\epsilon,n,\|\cdot\|_2
\right)
\le
r_wd
\log
\left(
1+\frac{2B_wB_x}{\epsilon}
\right).
\]
This completes the proof.
\end{proof}

Now, we recall Maurey's Sparsification Lemma.
\begin{lemma} \citep{Zhang2002NB}\label{lem:maurey} Let $H$ be a Hilbert space and let each $f \in \calH$ have the representation $f=\sum_{j=1}^d w_j g_j$, where each $\|g_j\|_2 \leq b, w_j \geq 0$ and $\alpha=\sum_{j=1}^d w_j \leq 1$. Then for every $t\geq 1$, there exists non-negative integers $k_1,k_2,\cdots,k_d\geq 0$ such that $\sum_{j=1}^d k_j \leq t$ and
\begin{align}
\bigg\|f-\frac{1}{t}\sum_{j=1}^d k_j g_j \bigg\|_2^2 \leq \frac{\alpha b^2-\|f\|_2^2}{t}.
\end{align}
\end{lemma}
Based on Lemma \ref{lem:maurey}, the following results can be drawn. 
\begin{theorem}
\label{thm:main1}
Let
\[
V_W\subseteq\mathbb{R}^k
\]
be a fixed linear subspace with
\[
\dim(V_W)=r_w.
\]
Define
\[
\mathcal{W}
=
\left\{
W\in\mathbb{R}^{k \times d}:
\mbox{col}(W) \subseteq V_W,\;
\|W \|_{2,1}\leq B_w
\right\},
\]
and
\[
\mathcal{F}^{(2,1)}
=
\left\{
f_W:x\mapsto Wx:
W\in\mathcal{W}
\right\}.
\]
Suppose that
\[
x_1,\ldots,x_N\in\mathbb{R}^d
\]
satisfy
\[
\|x_i\|_1 \leq B_x,
\qquad i=1,\ldots,n.
\]
Then, for every $\epsilon>0$,
\[
\log
\mathcal{N}_{\infty}
\left(
\mathcal{F}^{(2,1)},
\epsilon,N,\|\cdot\|_2
\right)
\lesssim
\frac{B_w^2B_x^2}{\epsilon^2}
\log(r_w d),
\]
where the implicit constant is universal.
\end{theorem}
This result improves upon \citep[Lemma 3.4]{Trauger2023LI} when $W$ is low-rank and $k\gg d$. A detailed proof of this theorem is provided in Section~\ref{thm:main1:proof}.

The following lemma is a result of combining Theorems \ref{thm:main0} and \ref{thm:main1}.
\begin{lemma}  \label{lem:aux2} For any $\eps \leq 2 B_x B_w$, we have
\begin{align}
\log
\mathcal{N}_{\infty}
\left(
\mathcal{F}^{(2,1)},
\epsilon,n,\|\cdot\|_2
\right)
\lesssim
\min\bigg\{\frac{r_wd}{2}\log \bigg(\frac{16 B_x^2 B_w^2  }{\eps^2} \bigg),\frac{B_w^2B_x^2}{\epsilon^2}
\log(r_w d) \bigg\} \label{askme}. 
\end{align}
\end{lemma}
\begin{remark}
In the special case $\mathcal{V}_W=\mathbb{R}^k$, the covering number bound in \eqref{askme} improves upon the covering number bound in \citep[Lemma 3.4]{Trauger2023LI} under the same norm constraints.
\end{remark}
\begin{proof}
Observe that $||W||_{2 \to 2} \leq B_w $ if $||W||_{2,1} \leq B_w$. In addition, $||x||_1\leq B_x$ then $||x||_2 \leq B_x$. Hence, we have $\calF^{(2,1)} \subset \calF^{(2,2)}$ under the same $B_x$ and $B_w$. It follows that
\begin{align}
\log
\mathcal{N}_{\infty}
\left(
\mathcal{F}^{(2,1)},
\epsilon,n,\|\cdot\|_2
\right)&= \min \bigg\{\log
\mathcal{N}_{\infty}
\left(
\mathcal{F}^{(2\to 2)},
\epsilon,n,\|\cdot\|_2
\right), \log
\mathcal{N}_{\infty}
\left(
\mathcal{F}^{(2,1)},
\epsilon,n,\|\cdot\|_2
\right)\bigg\} \nn\\
&\leq \min\bigg\{\frac{r_wd}{2}\log \bigg(\frac{16 B_x^2 B_w^2}{\eps^2} \bigg),\frac{B_w^2B_x^2}{\epsilon^2}
\log(r_w d) \bigg\} \label{eq9},
\end{align}
where \eqref{eq9} follows from the fact that
\[
r_wd
\log
\left(
1+\frac{2B_wB_x}{\epsilon}
\right) \leq \frac{r_wd}{2}\log \bigg(\frac{16 B_x^2 B_w^2}{\eps^2} \bigg)
\] whenever $\eps \leq 2 B_x B_w$. 
\end{proof}
Finally, we establish the following auxiliary result.
\begin{lemma}
\label{lem:small-epsilon-rank-entropy}
Assume that $r_wd>1$. Then there exists
\[
\epsilon_0
\in
\left(
0,\min\left\{2B_xB_w,\frac{B_x}{2}\right\}
\right]
\]
such that, for every
\[
0<\epsilon\leq\epsilon_0,
\]
we have
\begin{align}
\frac{r_wd}{2}
\log\left(
\frac{16B_x^2B_w^2}{\epsilon^2}
\right)
<
\frac{B_w^2B_x^2}{\epsilon^2}
\log(r_wd) \label{eq:small-epsilon-comparison}.
\end{align}
In particular, $\epsilon_0$ can be chosen so that
\[
\frac{r_wd\,\epsilon_0^2}{2B_x^2B_w^2}
\log\left(
\frac{16B_x^2B_w^2}{\epsilon_0^2}
\right)
<
\log(r_wd).
\]
\end{lemma}
\subsection{Covering Numbers under Rank Constraints}
Unlike the previous subsection, in this subsection we consider the case where the rank of $W$ is bounded, while the column space of $W$ is allowed to vary over arbitrary subspaces. We first establish the following result.
\begin{theorem}\label{thm:main0b}
Let
\[
\calW
=
\left\{
W\in\bbR^{k \times d}:
\|W\|_{2\to2}\leq B_w,\;
\operatorname{rank}(W)\leq r_w
\right\},
\]
where $1\leq r_w\leq \min\{d,k\}$, and let
\[
\calF_*^{(2\to 2)}
=
\left\{
f_W:x\mapsto Wx : W\in\calW
\right\}.
\]
Suppose that the inputs $x_1,\ldots,x_n\in\bbR^d$ satisfy
\begin{align}
\|x_i\|_2\leq B_x,\qquad i=1,\ldots,n.
\end{align}
Then, for every $0<\eps\leq B_xB_w$,
\begin{align}
\log
\calN_{\infty}
\left(
\calF_*^{(2\to 2)},\eps,n,\|\cdot\|_2
\right)
\leq
r_w(d+k-r_w)
\log\left(
\frac{C B_xB_w}{\eps}
\right),
\label{eq:covering-linear-spectral}
\end{align}
where $C>0$ is a universal constant independent of
$d,k,r_w,B_x,B_w,\eps$, and $n$.
\end{theorem}
\begin{remark}
By setting $r_w=k$, we obtain a novel covering number bound for the linear class
\[
\left\{x\mapsto Wx:\|W\|_{2\to 2}\leq B_w\right\}.
\]
By combining our bound with the existing covering number bounds of Edelman et al. \citep{Edelman2022} and Trauger and Tewari \citep{Trauger2023LI}, we further obtain improved covering number estimates under the same norm constraint (cf. Lemma \ref{lem:aux2b}).
\end{remark}

\begin{proof}
Let
\[
\delta=\frac{\eps}{B_x}.
\]
Since $0<\eps\leq B_xB_w$, we have
\[
0<\delta\leq B_w.
\]

For any $W,W'\in\calW$, we have
\begin{align}
\max_{1\leq i\leq n}
\|Wx_i-W'x_i\|_2
&=
\max_{1\leq i\leq n}
\|(W-W')x_i\|_2
\nonumber\\
&\leq
\max_{1\leq i\leq n}
\|W-W'\|_{2\to2}\|x_i\|_2
\nonumber\\
&\leq
B_x\|W-W'\|_{2\to2}.
\label{eq:linear-lipschitz}
\end{align}
Therefore, any $\delta$-cover of $\calW$ with respect to the
operator norm induces an $\eps$-cover of $\calF$. Hence,
\begin{align}
\calN_{\infty}
\left(
\calF_*^{(2\to 2)},\eps,n,\|\cdot\|_2
\right)
\leq
\calN
\left(
\calW,\delta,\|\cdot\|_{2\to2}
\right).
\label{eq:function-matrix-cover}
\end{align}

We next construct a $\delta$-cover of $\calW$.

For every $W\in\calW$, let
\[
s=\operatorname{rank}(W)\leq r_w.
\]
Consider a reduced singular value decomposition
\[
W=U_s\Sigma_sV_s^\top,
\]
where
\[
U_s\in\bbR^{d\times s},
\qquad
V_s\in\bbR^{k\times s},
\]
have orthonormal columns and
\[
\Sigma_s
=
\operatorname{diag}(\sigma_1,\ldots,\sigma_s),
\qquad
0<\sigma_j\leq B_w.
\]
Define the real Stiefel manifold by
\[
\operatorname{St}(m,r_w)
=
\left\{
Q\in\bbR^{m\times r_w}:Q^\top Q=I_{r_w}
\right\}.
\]
Extend $U_s$ and $V_s$ to matrices
\[
U\in\operatorname{St}(d,r_w),
\qquad
V\in\operatorname{St}(k,r_w),
\]
by adding orthonormal columns, and define
\[
\Sigma
=
\operatorname{diag}
(\sigma_1,\ldots,\sigma_s,
0,\ldots,0)
\in\bbR^{r_w\times r_w}.
\]
Then
\[
W=U\Sigma V^\top
\]
and
\[
\|\Sigma\|_{2\to2}\leq B_w.
\]

We use the following covering-number estimate for the Stiefel
manifold: there exists a universal constant $C_0>0$ such that, for
every $m\geq r_w$ and every $0<\eta<1$,
\begin{align}
\calN
\left(
\operatorname{St}(m,r_w),
\eta,
\|\cdot\|_{2\to2}
\right)
\leq
\left(
\frac{C_0}{\eta}
\right)^{
mr_w-\frac{r_w(r_w+1)}{2}
}.
\label{eq:stiefel-cover}
\end{align}
See, for example, Lemma 4.1 of
Hinrichs, Prochno, and Vybíral~\citep{HinrichsProchnoVybiral2017},
which derives this estimate from the metric entropy bounds for
homogeneous spaces developed by Szarek~\citep{Szarek1998}.

Choose
\[
\eta=\frac{\delta}{5B_w}.
\]
Since $\delta\leq B_w$, we have
\[
0<\eta\leq\frac{1}{5}<1.
\]
Hence, there exist finite sets
\[
\mathcal{U}\subset\operatorname{St}(d,r_w),
\qquad
\mathcal{V}\subset\operatorname{St}(k,r_w)
\]
such that
\begin{align}
|\mathcal{U}|
&\leq
\left(
\frac{C_0}{\eta}
\right)^{
dr_w-\frac{r_w(r_w+1)}{2}
},
\label{eq:U-cover}
\\
|\mathcal{V}|
&\leq
\left(
\frac{C_0}{\eta}
\right)^{
kr_w-\frac{r_w(r_w+1)}{2}
},
\label{eq:V-cover}
\end{align}
and for every
$U\in\operatorname{St}(d,r_w)$ and
$V\in\operatorname{St}(k,r_w)$, there exist
$\widehat U\in\mathcal{U}$ and
$\widehat V\in\mathcal{V}$ satisfying
\[
\|U-\widehat U\|_{2\to2}\leq\eta,
\qquad
\|V-\widehat V\|_{2\to2}\leq\eta.
\]

We next construct a cover for the singular values. Since $\Sigma$ is
diagonal,
\[
\|\Sigma-\widehat\Sigma\|_{2\to2}
=
\max_{1\leq j\leq r_w}
|\sigma_j-\widehat\sigma_j|.
\]
Let $\mathcal{S}$ be a uniform grid of $[0,B_w]^{r_w}$ with mesh
\[
\eta_\Sigma=\frac{\delta}{3}
\]
in the $\ell_\infty$ metric. Then, for every admissible $\Sigma$,
there exists $\widehat\Sigma\in\mathcal{S}$ such that
\[
\|\Sigma-\widehat\Sigma\|_{2\to2}
\leq
\eta_\Sigma.
\]
Furthermore,
\begin{align}
|\mathcal{S}|
&\leq
\left(
1+\frac{B_w}{\eta_\Sigma}
\right)^{r_w}
=
\left(
1+\frac{3B_w}{\delta}
\right)^{r_w}.
\label{eq:sigma-cover}
\end{align}

For any $W=U\Sigma V^\top\in\calW$, choose
$\widehat U\in\mathcal{U}$,
$\widehat V\in\mathcal{V}$, and
$\widehat\Sigma\in\mathcal{S}$ satisfying the above approximation
properties, and define
\[
\widehat W
=
\widehat U\widehat\Sigma\widehat V^\top.
\]
We have
\begin{align}
W-\widehat W
={}&
(U-\widehat U)\Sigma V^\top
+
\widehat U(\Sigma-\widehat\Sigma)V^\top
\nonumber\\
&+
\widehat U\widehat\Sigma
(V-\widehat V)^\top.
\label{eq:matrix-decomposition}
\end{align}
Therefore,
\begin{align}
\|W-\widehat W\|_{2\to2}
&\leq
\|U-\widehat U\|_{2\to2}
\|\Sigma\|_{2\to2}
\|V\|_{2\to2}
\nonumber\\
&\quad+
\|\widehat U\|_{2\to2}
\|\Sigma-\widehat\Sigma\|_{2\to2}
\|V\|_{2\to2}
\nonumber\\
&\quad+
\|\widehat U\|_{2\to2}
\|\widehat\Sigma\|_{2\to2}
\|V-\widehat V\|_{2\to2}.
\end{align}
Since
\[
\|U\|_{2\to2}
=
\|\widehat U\|_{2\to2}
=
\|V\|_{2\to2}
=1
\]
and
\[
\|\Sigma\|_{2\to2}\leq B_w,
\]
we obtain
\begin{align}
\|W-\widehat W\|_{2\to2}
&\leq
B_w\eta
+
\eta_\Sigma
+
\|\widehat\Sigma\|_{2\to2}\eta.
\label{eq:matrix-error}
\end{align}
Moreover,
\begin{align}
\|\widehat\Sigma\|_{2\to2}
&\leq
\|\Sigma\|_{2\to2}
+
\|\widehat\Sigma-\Sigma\|_{2\to2}
\nonumber\\
&\leq
B_w+\eta_\Sigma.
\end{align}
Thus,
\begin{align}
\|W-\widehat W\|_{2\to2}
&\leq
B_w\eta
+
\eta_\Sigma
+
(B_w+\eta_\Sigma)\eta.
\label{eq:matrix-error-final}
\end{align}

Substituting
\[
\eta=\frac{\delta}{5B_w},
\qquad
\eta_\Sigma=\frac{\delta}{3},
\]
and using $\delta\leq B_w$, we obtain
\begin{align}
\|W-\widehat W\|_{2\to2}
&\leq
\frac{\delta}{5}
+
\frac{\delta}{3}
+
\left(
B_w+\frac{\delta}{3}
\right)
\frac{\delta}{5B_w}
\nonumber\\
&\leq
\frac{\delta}{5}
+
\frac{\delta}{3}
+
\frac{4\delta}{15}
\nonumber\\
&=
\frac{4\delta}{5}
<\delta.
\end{align}
Therefore,
\[
\widehat{\calW}
=
\left\{
\widehat U\widehat\Sigma\widehat V^\top:
\widehat U\in\mathcal{U},\;
\widehat V\in\mathcal{V},\;
\widehat\Sigma\in\mathcal{S}
\right\}
\]
is a $\delta$-cover of $\calW$ under the operator norm. Hence,
\begin{align}
\calN
\left(
\calW,\delta,\|\cdot\|_{2\to2}
\right)
&\leq
|\mathcal{U}|
|\mathcal{V}|
|\mathcal{S}|
\nonumber\\
&\leq
\left(
\frac{C_0}{\eta}
\right)^{
dr_w-\frac{r_w(r_w+1)}{2}
}
\left(
\frac{C_0}{\eta}
\right)^{
kr_w-\frac{r_w(r_w+1)}{2}
}
\left(
1+\frac{3B_w}{\delta}
\right)^{r_w}.
\end{align}

Substituting
\[
\eta=\frac{\delta}{5B_w}
\]
gives
\begin{align}
\calN
\left(
\calW,\delta,\|\cdot\|_{2\to2}
\right)
&\leq
\left(
\frac{5C_0B_w}{\delta}
\right)^{
dr_w-\frac{r_w(r_w+1)}{2}
}
\left(
\frac{5C_0B_w}{\delta}
\right)^{
kr_w-\frac{r_w(r_w+1)}{2}
}
\left(
1+\frac{3B_w}{\delta}
\right)^{r_w}.
\end{align}
Since $\delta\leq B_w$,
\[
1+\frac{3B_w}{\delta}
\leq
\frac{4B_w}{\delta}.
\]
Therefore, for a universal constant $C_1>0$,
\begin{align}
\log
\calN
\left(
\calW,\delta,\|\cdot\|_{2\to2}
\right)
&\leq
\left[
dr_w-\frac{r_w(r_w+1)}{2}
+
kr_w-\frac{r_w(r_w+1)}{2}
+
r_w
\right]
\log\left(
\frac{C_1B_w}{\delta}
\right)
\nonumber\\
&=
r_w(d+k-r_w)
\log\left(
\frac{C_1B_w}{\delta}
\right).
\label{eq:matrix-cover}
\end{align}

Finally, substituting
\[
\delta=\frac{\eps}{B_x}
\]
into \eqref{eq:function-matrix-cover} gives
\begin{align}
\log
\calN_{\infty}
\left(
\calF_*^{(2\to 2)},\eps,n,\|\cdot\|_2
\right)
&\leq
r_w(d+k-r_w)
\log\left(
\frac{C_1B_xB_w}{\eps}
\right).
\end{align}
Renaming $C_1$ as $C$ completes the proof.
\end{proof}
The following lemma shows that under the same condition as Lemma 3.6 in \citep{Trauger2023LI}, we can achieve a stricter bound by using Theorem \ref{thm:main0b}. 
\begin{lemma}  \label{lem:aux2b} For any $\eps>0$, we have
\begin{align}
\log
\mathcal{N}_{\infty}
\left(
\mathcal{F}_*^{(2,1)},
\epsilon,n,\|\cdot\|_2
\right)
\lesssim
\min\bigg\{\frac{r_w(d+k-r_w)}{2}
\log\left(
\frac{C^2B_x^2B_w^2}{\eps^2}
\right),\frac{B_w^2B_x^2}{\epsilon^2}
\log(k d) \bigg\} \label{askmeb}
\end{align} for some universal constant $C>0$, which is independent of $d, k, r_w, B_x, B_w, \eps,$ and $n$. 
\end{lemma}
\begin{remark}
In the special case $r_w=k$, the covering number bound in \eqref{askmeb} improves upon the covering number bound in \citep[Lemma 3.4]{Trauger2023LI} under the same norm constraints.
\end{remark}
\begin{proof}
Let 
\[
\calF_*^{(2,1)}=\{x\mapsto Wx: W \in \calW_*, ||x||_1 \leq 1\},
\]
where
\begin{align}
\calW_*=\{W \in \bbR^{k \times d}: \|W\|_{2,1} \leq B_w,  \mbox{rank}(W) \leq r_w\}.
\end{align}
Observe that $||W||_{2 \to 2} \leq B_w $ if $||W||_{2,1} \leq B_W$. In addition, $||x||_1\leq B_x$ then $||x||_2 \leq B_x$. Hence, we have $\calF_*^{(2,1)} \subset \calF_*^{(2\to 2)}$ under the same $B_x$ and $B_w$. It follows that
\begin{align}
\log
\mathcal{N}_{\infty}
\left(
\mathcal{F}_*^{(2,1)},
\epsilon,n,\|\cdot\|_2
\right)&= \min \bigg\{\log
\mathcal{N}_{\infty}
\left(
\mathcal{F}_*^{(2\to 2)},
\epsilon,n,\|\cdot\|_2
\right), \log
\mathcal{N}_{\infty}
\left(
\mathcal{F}_*^{(2,1)},
\epsilon,N,\|\cdot\|_2
\right)\bigg\} \nn\\
&\lesssim \min\bigg\{\frac{r_w(d+k-r_w)}{2}
\log\left(
\frac{C^2B_x^2B_w^2}{\eps^2}
\right),\frac{B_w^2B_x^2}{\epsilon^2}
\log(k d) \bigg\} \label{amot}.
\end{align}
Here, \eqref{amot} follows from Theorem \ref{thm:main0b} and the fact that
\begin{align}
\log
\mathcal{N}_{\infty}
\left(
\mathcal{F}_*^{(2,1)},
\epsilon,N,\|\cdot\|_2 
\right) \leq  \log
\mathcal{N}_{\infty}
\left(
\calG,
\epsilon,N,\|\cdot\|_2\right),
\end{align}
where
\begin{align}
\calG=\{x \mapsto Wx: \|W\|_{2\to 2} \leq B_w\}.
\end{align}
Note that by \citep[Lemma 3.4]{Trauger2023LI}, 
\begin{align}
\log
\mathcal{N}_{\infty}
\left(
\calG,
\epsilon,N,\|\cdot\|_2\right) \lesssim \frac{B_w^2B_x^2}{\epsilon^2}
\log(k d). 
\end{align}
\end{proof}
Finally, similar to the previous subsection, we establish the following auxiliary result. 
\begin{lemma}
\label{lem:small-epsilon-rank-entropyb}
Assume that $kd>1$. Then there exists
\[
\epsilon_0
\in
\left(
0, \frac{B_x}{2}
\right]
\]
such that, for every
\[
0<\epsilon\leq\epsilon_0,
\]
we have
\begin{align}
\frac{r_w(d+k-r_w)}{2}
\log\left(
\frac{C^2B_x^2B_w^2}{\eps^2}\right)
<
\frac{B_w^2B_x^2}{\epsilon^2}
\log(k d)  \label{eq:small-epsilon-comparisonb}.
\end{align}
In particular, $\epsilon_0$ can be chosen so that
\[
\frac{r_w(d+k-r_w)\eps_0^2}{2B_w^2 B_x^2}
\log\left(
\frac{C^2B_x^2B_w^2}{\eps_0^2}\right)
<
\log(k d).
\]
\end{lemma}

\section{Softmax Sensitivity}
\label{sec:softmax-sensitivity}

The attention mechanism contains a nonlinear row-wise softmax operation,
and it is therefore important to understand how sensitive the attention
weights are to perturbations of the parameter matrix $W_{QK}$. Recall
that
\[
\operatorname{Att}_{W_{QK},W_v,W_c}(X)
=
\sigma\!\left(
\operatorname{RowSoftmax}
\left(
XW_{QK}X^\top
\right)
XW_v
\right)W_c.
\]

Define the attention score matrix
\[
S(W_{QK}) = XW_{QK}X^\top
\]
and the corresponding attention matrix
\[
A(W_{QK})
=
\operatorname{RowSoftmax}(S(W_{QK})).
\]
The row-wise softmax map is Lipschitz. In particular, there exists a
universal constant $L_{\mathrm{sm}}>0$ such that
\begin{align}
\left\|
\operatorname{RowSoftmax}(S)
-
\operatorname{RowSoftmax}(S')
\right\|_F
\leq
L_{\mathrm{sm}}
\|S-S'\|_F.
\label{eq:softmax-lipschitz}
\end{align}
Therefore, for two matrices $W_{QK}$ and $W_{QK}'$,
\begin{align}
\|A(W_{QK})-A(W_{QK}')\|_F
&\leq
L_{\mathrm{sm}}
\left\|
X(W_{QK}-W_{QK}')X^\top
\right\|_F
\nonumber\\
&\leq
L_{\mathrm{sm}}
\|X\|_{2\to 2}^2
\|W_{QK}-W_{QK}'\|_F.
\label{eq:softmax-qk-sensitivity}
\end{align}
Thus, on a bounded input set, a perturbation of $W_{QK}$ produces a
controlled perturbation of the attention weights.

The effect of this perturbation on the attention output can be made
explicit. Let
\[
H(W_{QK},W_v)
=
A(W_{QK})XW_v.
\]
Then
\begin{align}
&H(W_{QK},W_v)
-
H(W_{QK}',W_v')
\nonumber\\
&=
\left[A(W_{QK})-A(W_{QK}')\right]XW_v
+
A(W_{QK}')X(W_v-W_v').
\label{eq:softmax-output-decomposition}
\end{align}
Consequently,
\begin{align}
\|H(W_{QK},W_v)-H(W_{QK}',W_v')\|_F
\leq
C_1\|W_{QK}-W_{QK}'\|_F
+
C_2\|W_v-W_v'\|_F,
\label{eq:softmax-output-sensitivity}
\end{align}
where $C_1$ and $C_2$ depend on bounds on $X$, $W_v$, and the
softmax Lipschitz constant $L_{\mathrm{sm}}$.

Since $\sigma$ is assumed to be $L_\sigma$-Lipschitz with
$\sigma(0)=0$, and the final operation is multiplication by $W_c$,
the same perturbation analysis can be propagated through the remaining
components of the attention head. In particular, under suitable norm
bounds on $W_c$, there exists a constant $C_{\mathrm{Att}}>0$ such that
\begin{align}
\left\|
\operatorname{Att}_{W_{QK},W_v,W_c}(X)
-
\operatorname{Att}_{W_{QK}',W_v',W_c'}(X)
\right\|_F
\leq
C_{\mathrm{Att}}
\left(
\|W_{QK}-W_{QK}'\|_F
+
\|W_v-W_v'\|_F
+
\|W_c-W_c'\|_F
\right).
\label{eq:attention-softmax-sensitivity}
\end{align}

Equation~\eqref{eq:attention-softmax-sensitivity} shows that the
softmax operation does not invalidate the covering-number analysis
based on the underlying linear parameterizations. Instead, its effect
is captured through a Lipschitz sensitivity factor. In particular, the
softmax does not by itself imply an additional rank-dependent entropy
term; any rank dependence in the covering number arises from the
parameter classes and the corresponding covering construction.

For the scalar prediction setting considered in this work, where
\[
f(X)=w^\top Y_{\texttt{[CLS]}},
\]
the sensitivity of the final prediction can likewise be controlled by
the perturbation of the final representation. Hence, provided the
intermediate parameter norms are bounded, the softmax contributes only
multiplicative sensitivity constants to the resulting generalization
and covering-number bounds.

This analysis indicates that the nonlinear softmax transformation is
compatible with the rank-dependent covering framework: parameter
perturbations are first propagated through the score matrix
$XW_{QK}X^\top$, then through the Lipschitz softmax map, and finally
through the remaining Lipschitz components of the attention head.

\section{Rank-Dependent Generalization Error Bounds for Single Layer Transformer}
\subsection{Single Layer Single Head} \label{sub:1}
Let \( w \in \mathbb{R}^d \), \( W_c \in \mathbb{R}^{k \times d} \), \( W_v \in \mathbb{R}^{d \times k} \), and \( W_{QK} \in \mathbb{R}^{d \times d} \). Following \citep{Trauger2023LI}, we define the scalar output of a one-layer Transformer as \( w^\top Y_{\texttt{[CLS]}} \), where
\[
Y_{\texttt{[CLS]}} = W_c^\top \, \sigma\left(W_v^\top X^\top \, \mathrm{softmax}(X W_{QK}^\top x_{\texttt{[CLS]}})\right),
\]
and \( \sigma \) denotes a nonlinearity applied element-wise.

The Rademacher complexity associated with this architecture is given by \citep{Trauger2023LI}
\[
\bbE\big[\sup_{w, W_c \in \calW_c, W_v \in \calW_v, W_{QK}\in \calW_{QK}} \sum_{i=1}^n \epsilon_i \, W_c^\top  \sigma\big(W_v^\top X_{(i)}^\top \, \mathrm{softmax}(X_{(i)} W_{QK}^\top x_{\texttt{[CLS]}})\big)\big],
\]
where \( \{X_{(i)}\}_{i=1}^n \), with \( X_{(i)} \in \mathbb{R}^{T \times d} \), are the input data samples, and \( \epsilon_i \) are i.i.d.\ Rademacher random variables taking values in \( \{-1, +1\} \) with equal probability.

Now, we recall Lemma 27.4 in \citep{ShalevShwartz2014UnderstandingML}.
\begin{lemma}\citep{ShalevShwartz2014UnderstandingML} \label{lem:dudley} Let $\calF=\{f: \calX \to \bbR\}$. For any $\{x_1,x_2,\cdots,x_n\} \in \calX^n$ and $\sup_{i \in [n]} \sup_{f \in \calF}|f(x_i)| \leq c_x$ we have
\begin{align}
&\calR_n(\calF,\{x_i\}_{i=1}^n) \leq 2 \eps_{m+1}\nn\\
& + \frac{12}{\sqrt{n}} \sum_{j=1}^m (\eps_j-\eps_{j+1}) \sqrt{\log N_{\infty}(\calF, \eps_j,  \|\cdot\|_2)}, \enspace \forall m \in \bbZ_+, 
\end{align}
where $\eps_j:= c_x/2^j$ for all $j \in [m+1]$. 
\end{lemma}
By applying Lemma~\ref{lem:dudley} together with the novel log-covering number bounds established in Section~\ref{sec:theory}, we obtain a Rademacher complexity bound for transformers that depends on the rank $r_w$ of the weight matrices and decays as $O(1/\sqrt{n})$ in the sample size. This improves upon Corollary~4.1.1 of \citep{Trauger2023LI} and Lemma~A.5 of \citep{Edelman2022}, which yield bounds of order $O\!\left(\frac{\log n}{\sqrt{n}}\right)$. We summarize this result in the following theorem.
\begin{theorem} \label{thm:main}
Suppose that $\log \calN_{\infty}(\calH, \frac{\eps}{2B_x^2}, n, \|\cdot\|_2)$ for the function class $\calH:=\{x \to W_{QK} x: x\in \calX, W_{QK} \in \calW\}$ is bounded by $\min\big\{a\log \big(\frac{b}{\eps^2}\big), \frac{q^2}{\eps^2}\big\}$ where $a\log \big(\frac{b}{\eps^2}\big)\leq \frac{q^2}{\eps^2}$ for all $\eps \leq \eps_0$ for some $\eps_0 \leq B_x/2$.  In addition,  $\|w\|_1 \leq B_w, \|W_c^T\|_{1,\infty} \leq B_{W_c}$, $\|W_v^T\|_{1,\infty}\leq B_{W_v}$, and $\|X_{(i)}\|_{1,\infty} \leq B_x$.  
Then, the Rademacher complexity satisfies
\begin{align}
&\calR_n(\calF, \{X_{(i)}\}_{i=1}^n ) \nn\\
&:= \bbE\bigg[\sup_{w,W_c\in \calW_c,W_v \in \calW_v,W_{QK} \in \calW_{QK}} \sum_{i=1}^n \sigma_i W_c^T \sigma(W_v^T X_{(i)}^T
\mbox{softmax}(X_{(i)} W_{QK}^T x_{[CLS]}))\bigg]\nn\\
&  \leq \frac{24B_w B_{W_c}L_{\sigma}B_{W_c}}{\sqrt{n}}\bigg( \eps_0 \bigg[ \sqrt{a \log \bigg(\frac{b}{B_x^2}\bigg)}\nn\\
& + \sqrt{a \log 4} \bigg(\bigg\lceil \log_2 \frac{B_x}{\eps_0}\bigg \rceil+1  \bigg)\bigg]+ q \log \bigg(\frac{B_x}{\eps_0}\bigg)\bigg). 
\end{align}
\end{theorem}
\begin{proof}
By using Talagrand's contraction lemma \citep{LedouxT1991book} and properties of softmax activation function (cf. Corollary A.7 in \citep{Edelman2022}), it is shown that (cf. Proof of Theorem 4.1 in \citep{Trauger2023LI}):
\begin{align}
R_n(\calF, \{X_{(i)}\}_{i=1}^n) \leq 2B_w B_{W_c}L_{\sigma}B_{W_c} \Lambda    \label{matge0},
\end{align}
where
\begin{align}
\Lambda&:=\bbE\bigg[\sup_{s,j \in [d]} \sup_{W_{QK}} \sum_{i=1}^m s \sigma_i e_j^T X_{(i)}^T  \mbox{softmax}(X_{(i)} W_{QK} x_{[CLS]})  \bigg]. 
\end{align}
Now, as \citep{Trauger2023LI}, we bound $\Lambda$ by considering the following function class:
\begin{align*}
\calS&=\big\{X \to s e_j^T X^T \mbox{softmax}(XW_{QK} x_{[CLS]}): \nn\\
&\qquad  s \in \{-1,+1\}, j \in [d], W_{QK} \in \calW\big\}. 
\end{align*}
Let $\calH=\{x \to W_{QK}x\}$ be a class of linear functions and $\hat{\calH}$ be an $\frac{\eps}{2B_x^2}$-cover for this function class.  Then, for any $W_{QK} \in \calW$ and $x_{[CLS]}\in \bbR^d$, there exists $\hat{V}_{QK} \in \hat{\calH}$ such that
\begin{align*}
\big\|W_{QK} x_{[CLS]}- \hat{V}_{QK} \big\| \leq \frac{\eps}{2B_x^2}  \label{matge2}.
\end{align*}

Hence, by the proof of Theorem 4.1 in \citep{Trauger2023LI}, we have
\begin{align}
&\bigg\|s e_j^T X^T \mbox{softmax}(XW_{QK} x_{[CLS]}) 
-s e_j^T X^T \mbox{softmax}(X\hat{V}_{QK} )\bigg\|\nn\\
&\qquad \leq 2 B_x^2 \big\|W_{QK} x_{[CLS]}- \hat{V}_{QK} \big\| \nn \\
&\qquad \leq \eps.
\end{align}

Furthermore, by our assumption we also have
\begin{align*}
&\log |\hat{\calH}|=\log N_{\infty}\bigg(\calH, \frac{\eps}{2B_x^2},n,\|\cdot\|_2\bigg) \nn\\
&\qquad \leq \min\bigg\{a \log \bigg(\frac{b}{\eps^2}\bigg), \frac{q^2}{\eps^2}\bigg)\bigg\} . 
\end{align*}
It follows that
\begin{align}
\log N_{\infty}\bigg(\calS, \eps,n,\|\cdot\|_2\bigg) \leq \min\bigg\{a \log \bigg(\frac{b}{\eps^2}\bigg), \frac{q^2}{\eps^2}\bigg)\bigg\} \label{tfact2}.
\end{align}
Also, due to the sofmax in $\calS$, the largest value the function class can be is $B_x$. 

Now, let
\[
\eps_j=\frac{B_x}{2^j}, \qquad \forall j \in \bbZ+,
\]
Let $m_0=\big\lceil \log_2 \frac{B_x}{\eps_0}\big \rceil$. Then, we also have
\[
\eps_{m_0-1}=\frac{B_x}{2^{m_0-1}}\geq \eps_0\geq \frac{B_x}{2^{m_0}}=\eps_{m_0}. 
\]
Then, for any $m \in \bbZ_+$ by Lemma \ref{lem:dudley},  it holds that
\begin{align}
\Lambda
&\leq 2 \eps_{m+1} + \frac{12}{\sqrt{n}} \sum_{j=1}^m (\eps_j-\eps_{j+1})\nn\\
&\qquad \times \min\bigg\{\sqrt{a \log (b/B_x^2)+ j  a\log 4}, \frac{q}{\eps_j}\bigg\} \nn \\
&= 2 \eps_{m+1} + \frac{12}{\sqrt{n}} \sum_{j=m_0}^m (\eps_j-\eps_{j+1})\nn\\
&\qquad \times \min\bigg\{\sqrt{a \log (b/B_x^2)+ j a \log 4}, \frac{q}{\eps_j}\bigg\}\nn\\
&+ \frac{12}{\sqrt{n}} \sum_{j=1}^{m_0-1} (\eps_j-\eps_{j+1})\nn\\
&\qquad \times \min\bigg\{\sqrt{a \log (b/B_x^2)+ j a \log 4}, \frac{q}{\eps_j}\bigg\} \nn \\
&\leq 2 \eps_{m+1} + \frac{12}{\sqrt{n}} \sum_{j=m_0}^m (\eps_j-\eps_{j+1}) \nn\\
&\qquad \times \sqrt{a \log (b/B_x^2)+ j a \log 4}\nn\\
&\qquad + \frac{12}{\sqrt{n}} \sum_{j=1}^{m_0-1} (\eps_j-\eps_{j+1}) \frac{q}{\eps_j}  \label{eq65} \\
&\leq 2 \eps_{m+1} + \frac{12}{\sqrt{n}} \sum_{j=m_0}^m (\eps_j-\eps_{j+1})\nn\\
&\times  \sqrt{a \log (b/B_x^2)+ j a \log 4}+ \frac{12}{\sqrt{n}} \int_{\eps_{m_0}}^{B_x/2} \frac{q}{\eps} d\eps. 
\end{align}
Note that \eqref{eq65} is tighter than using only the assumption that the log covering number for the function class $\{x \to W_{QK} x: x\in \calX, W_{QK} \in \calW\}$  is bounded by $ \frac{q^2}{\eps^2}$ (cf. Theorem 4.1 in \citep{Trauger2023LI}) since $a\log \big(\frac{b}{\eps^2}\big)\leq \frac{q^2}{\eps^2}$ for all $\eps \leq \eps_0$. 

Now, we have
\begin{align}
&\sum_{m_0}^m (\eps_j-\eps_{j+1}) \sqrt{a \log (b/B_x^2)+ j a \log 4} \nn\\
&\qquad \leq \sum_{j=m_0}^m (\eps_j-\eps_{j+1})\sqrt{a \log (b/B_x^2)}\nn\\
&\qquad \qquad +  \sum_{j=m_0}^m (\eps_j-\eps_{j+1})\sqrt{a \log 4}\sqrt{j} \nn \\
&\qquad \leq (\eps_{m_0}-\eps_{m+1})\sqrt{a \log (b/B_x^2)}\nn\\
&\qquad \qquad + \sqrt{ a\log 4}\sum_{j=m_0}^m (\eps_j-\eps_{j+1})j \nn.
\end{align}
On the other hand, we have
\begin{align}
&\sum_{j=m_0}^m (\eps_j-\eps_{j+1})j\nn\\
&\qquad =B_x \sum_{j=m_0}^m j 2^{-(j+1)} \nn \\
&\qquad =2B_x \sum_{j=m_0+1}^{m+1} (j-1) 2^{-(j+1)} \nn \\
&\qquad =2B_x \sum_{j=m_0+1}^{m+1} \big(j 2^{-j}- (j+1) 2^{-(j+1)}\big) \nn \\
&\qquad = 2B_x \big((m_0+1) 2^{-(m_0+1)} - (m+2) 2^{-(m+2)}\big). 
\end{align}
Combine the above facts, we have
\begin{align}
\Lambda &\leq 2 \eps_{m+1} + \frac{12}{\sqrt{n}} \bigg((\eps_{m_0}-\eps_{m+1})\sqrt{a \log (b/B_x^2)}\nn\\
&+ 2B_x\sqrt{ a\log 4} \big((m_0+1) 2^{-(m_0+1)} - (m+2) 2^{-(m+2)}\big)\bigg)\nn\\
& + \frac{12}{\sqrt{n}} \int_{\eps_{m_0}}^{B_x/2} \frac{q}{\eps} d\eps. 
\end{align}

Let $m\to \infty$, we finally have
\begin{align}
\Lambda &\leq \frac{12}{\sqrt{n}} \bigg(\eps_{m_0} \sqrt{a \log (b/B_x^2)}\nn\\
& \qquad + 2B_x\sqrt{a \log 4} \big((m_0+1) 2^{-(m_0+1)} \big)\bigg)\nn\\
&\qquad + \frac{12}{\sqrt{n}} \int_{\eps_{m_0}}^{B_x/2} \frac{q}{\eps} d\eps\\
&=\frac{12}{\sqrt{n}} \bigg(\eps_{m_0} \sqrt{a \log (b/B_x^2)}\nn\\
& + 2B_x \sqrt{a\log 4} \big((m_0+1) 2^{-(m_0+1)} \big)\bigg)\nn\\
&\qquad + \frac{12q}{\sqrt{n}} \log \bigg(\frac{B_x}{2\eps_{m_0}}\bigg) \label{tagema}.
\end{align}
Now, observe that
\begin{align}
\frac{B_x}{2\eps_{m_0}}&=\frac{B_x}{\eps_{m_0-1}}\leq \frac{B_x}{\eps_0}, \nn \\
\eps_{m_0}&\leq \eps_0, \nn \\
2^{-(m_0+1)}&= \frac{\eps_{m_0}}{2B_x} \leq \frac{\eps_0}{2B_x}. \nn 
\end{align}
Hence, from \eqref{tagema} we obtain
\begin{align}
\Lambda & \leq \frac{12\eps_0}{\sqrt{n}}\bigg[ \sqrt{a \log \bigg(\frac{b}{B_x^2}\bigg)}\nn\\
&+ \sqrt{a \log 4} \bigg(\bigg\lceil \log_2 \frac{B_x}{\eps_0}\bigg \rceil+1  \bigg)\bigg]+ \frac{12q}{\sqrt{n}} \log \bigg(\frac{B_x}{\eps_0}\bigg) \label{xime2}.
\end{align}
From \eqref{matge0} and \eqref{xime2} we have
\begin{align}
&R_n(\calF, \{X_{(i)}\}_{i=1}^n) \nn\\
&\leq \frac{24B_w B_{W_c}L_{\sigma}B_{W_c}}{\sqrt{n}}\bigg( \eps_0 \bigg[ \sqrt{a \log \bigg(\frac{b}{B_x^2}\bigg)}\nn\\
&\qquad + \sqrt{ a\log 4} \bigg(\bigg\lceil \log_2 \frac{B_x}{\eps_0}\bigg \rceil+1  \bigg)\bigg]+ q \log \bigg(\frac{B_x}{\eps_0}\bigg)\bigg) \nn. 
\end{align}
\end{proof}
\begin{corollary}  \label{cor:18} Under the condition that $W_{QK} \in \calW_{QK}:=\{W: \|W\|_{2,1} \leq B_w, W \in \calV_W\}$ for some $V_W \in \bbR^d$ with $\dim(V_W)=r_w$, and $\|X_{(i)}\|_{1, \infty} \leq B_x$, the Rademacher complexity satisfies
 \begin{align}
&R_n(\calF, \{X_{(i)}\}_{i=1}^n) \nn\\
&\leq \frac{24B_w B_{W_c}L_{\sigma}B_{W_c}}{\sqrt{n}}\bigg( \eps_0 \bigg[ \sqrt{a \log \bigg(\frac{b}{B_x^2}\bigg)}\nn\\
&\qquad +\sqrt{ a\log 4} \bigg(\bigg\lceil \log_2 \frac{B_x}{\eps_0}\bigg \rceil+1  \bigg)\bigg]+ q \log \bigg(\frac{B_x}{\eps_0}\bigg)\bigg),
\end{align}
where
\begin{align}
a=\frac{r_wd}{2},\quad b= 16 B_x^2 B_w^2, \quad q= B_w B_x\sqrt{\log(r_w d)}
\end{align} and $\eps_0$ is chosen according to Lemma \ref{lem:small-epsilon-rank-entropy}. 
\end{corollary}
\begin{corollary}  \label{cor:18b} Under the condition that $W_{QK} \in \calW_{QK}:=\{W: \|W\|_{2,1} \leq B_w, \mbox{rank}(W)\leq r_w\}$, and $\|X_{(i)}\|_{1,\infty} \leq B_x$, the Rademacher complexity satisfies
 \begin{align}
&R_n(\calF, \{X_{(i)}\}_{i=1}^n) \nn\\
&\leq \frac{24B_w B_{W_c}L_{\sigma}B_{W_c}}{\sqrt{n}}\bigg( \eps_0 \bigg[ \sqrt{a \log \bigg(\frac{b}{B_x^2}\bigg)}\nn\\
&\qquad + \sqrt{ a\log 4} \bigg(\bigg\lceil \log_2 \frac{B_x}{\eps_0}\bigg \rceil+1  \bigg)\bigg]+ q \log \bigg(\frac{B_x}{\eps_0}\bigg)\bigg),
\end{align}
where
\begin{align}
a=\frac{r_wd}{2},\quad b= C^2  B_x^2 B_w^2, \quad q= B_w B_x\sqrt{\log(k d)}
\end{align} and $\eps_0$ is chosen according to Lemma \ref{lem:small-epsilon-rank-entropyb}. 
\end{corollary}
\begin{remark} Some remarks:
\begin{itemize}
\item Our Rademacher complexity bounds in Corollaries~\ref{cor:18} and~\ref{cor:18b} improve upon Corollary~4.1.1 of \citep{Trauger2023LI} by taking the minimum of two complementary covering-number bounds. Moreover, our analysis explicitly captures the dependence on the rank $r_w$, which yields a sharper characterization of the complexity for low-rank hypothesis classes.
\item Although trained Transformer weight matrices are typically full rank in
floating-point arithmetic, their singular values may exhibit substantial
decay. Thus, our rank-dependent results can also be interpreted in terms of
an accuracy-dependent effective rank. In particular, if $W_{\leq r}$ denotes
the rank-$r$ spectral truncation of $W$, then
\[
\|W-W_{\leq r}\|_{2\to 2}=\sigma_{r+1}(W),
\]
so for $\|x\|_2\leq B_x$ the truncation error is at most
$B_x\sigma_{r+1}(W)$. Hence, one may choose $r$ according to the desired
accuracy and apply our rank-dependent bound to $W_{\leq r}$. This provides
an approximate-rank interpretation of our results beyond the exact-rank
setting.

This interpretation is particularly relevant for layers whose singular
values decay rapidly, so that a relatively small number of singular
components captures most of the action of the corresponding linear
transformation. In this case, the effective rank can be substantially
smaller than the ambient dimension, making the rank-dependent bounds more
informative. The same interpretation applies to the query-key and
value-related matrices, where the effective rank characterizes the
dimensionality of the corresponding attention representations. We emphasize
that small effective rank is not assumed to hold universally for trained
Transformers; rather, our bounds are most informative for models or layers
that exhibit such spectral decay.

\item For all other parameters fixed, 
$\mathcal{R}_n(\mathcal{F}, \{X_{(i)}\}_{i=1}^n)=O(1/\sqrt{n})$,
which improves upon Corollary~4.1.1 of \citep{Trauger2023LI} and Lemma~A.5 of \citep{Edelman2022}, where the bounds scale as 
$O\!\left(\frac{\log n}{\sqrt{n}}\right)$.
This improvement follows from our use of the following sharper bound on the log-covering number:
\begin{align*}
\log N_{\infty}\!\left(\calH, \varepsilon, n, \|\cdot\|_2\right)
\leq
\min\!\left\{
a \log\!\left(\frac{b}{\varepsilon^2}\right),
\frac{q^2}{\varepsilon^2}
\right\},
\end{align*} which is developed in Section \ref{sec:theory}. 

This bound is tighter than the one obtained by assuming
\[
\log N_{\infty}\!\left(\calH, \varepsilon, n, \|\cdot\|_2\right)
= O\!\left(\frac{q^2}{\varepsilon^2}\right),
\]
which corresponds to applying Theorem~4.1 of \citep{Trauger2023LI}.
\end{itemize}
\end{remark}

\subsection{Single Layer Multiple Heads}
As shown in Section~4.2 in \citep{Trauger2023LI}, the Rademacher complexity of a
single-layer Transformer exhibits linear dependence on the number of attention
heads $H$. Extensions to multi-layer architectures follow by applying the same
reasoning as in Section~4.3 in \citep{Trauger2023LI}.

\subsection{Extension to Multi-layer Transformers and Vector-valued Outputs}
\label{sec:extension}

Our analysis focuses on a single-layer Transformer with a scalar output,
which allows us to isolate the effect of the rank-dependent constraints
on the covering number. The same covering-number arguments can serve as
building blocks for analyzing multi-layer Transformer architectures.
Consider a Transformer consisting of $L$ layers, and let $\mathcal{F}_\ell$
denote the function class associated with the $\ell$-th layer. If the
individual layers are Lipschitz continuous, covering the composition of
these function classes can be achieved by successively applying the
covering bounds for the individual layers. Consequently, the
rank-dependent terms associated with the individual layers can propagate
through the network, with their contribution controlled by the
layer-wise Lipschitz constants.

More specifically, suppose that the $\ell$-th layer has Lipschitz
constant $L_\ell$ and covering radius $\epsilon_\ell$. An approximation
error introduced at layer $\ell$ can be propagated through the subsequent
layers, resulting in a contribution proportional to
\[
\epsilon_\ell \prod_{j=\ell+1}^{L} L_j.
\]
Thus, by choosing the layer-wise covering radii such that
\[
\sum_{\ell=1}^{L}
\epsilon_\ell
\prod_{j=\ell+1}^{L} L_j
\leq \epsilon,
\]
one can obtain an $\epsilon$-cover of the overall multi-layer function
class. The corresponding metric entropy can then be bounded by combining
the covering numbers of the individual layers. This indicates that the
rank-dependent complexity terms derived in this paper can be propagated
through multiple layers, although the resulting bound depends on the
layer-wise Lipschitz constants and the ranks of the individual layers.

Residual connections can be incorporated into this framework as well.
Indeed, a residual mapping has the form
\[
x \mapsto x + f_\ell(x),
\]
and its Lipschitz constant satisfies
\[
\operatorname{Lip}(I+f_\ell)
\leq 1+\operatorname{Lip}(f_\ell).
\]
Hence, the identity mapping introduced by a residual connection does not
introduce an additional rank-dependent complexity term; rather, it
affects the propagation of approximation errors through its Lipschitz
constant. The rank-dependent covering bounds for the nonlinear component
$f_\ell$ can therefore be combined with the standard composition
argument.

Our scalar-output analysis can also be extended to vector-valued
outputs. Let
$f:\mathcal{X}\rightarrow\mathbb{R}^{m}$ and suppose that the output is
measured under the Euclidean norm. A vector-valued cover can be obtained
by controlling the approximation error jointly across the $m$ output
coordinates. In particular, a coordinate-wise construction yields
\[
\log \mathcal{N}(\epsilon,\mathcal{F},\|\cdot\|_2)
\leq
\sum_{j=1}^{m}
\log \mathcal{N}
\left(
\frac{\epsilon}{\sqrt{m}},
\mathcal{F}_j,\|\cdot\|_2
\right),
\]
where $\mathcal{F}_j$ denotes the function class corresponding to the
$j$-th output coordinate. Thus, the scalar-output covering bounds
developed in this paper can be used as a building block for obtaining
vector-valued bounds, with an additional dependence on the output
dimension.

A complete analysis of deep Transformers would require a careful
layer-wise treatment of the attention, feed-forward, normalization, and
residual components, together with their respective Lipschitz constants.
We leave such a full multi-layer analysis for future work. Nevertheless,
the above discussion shows that the rank-dependent covering arguments
developed here are not restricted conceptually to a single-layer
architecture and provide a basis for extending the analysis to deeper
Transformer networks.

\section{Comparison with existing research results}
In this section, we compare the results of \citep{Edelman2022} and \citep{Trauger2023LI} with our results in the case where only norm constraints are imposed. Our covering number bounds under norm-only constraints are obtained by setting $\mathcal{V}_W=\mathbb{R}^k$ or $r_w=k$ in the corresponding theorems and lemmas. Hence, our covering number and generalization bounds also apply to the norm-only setting considered in \citep{Edelman2022} and \citep{Trauger2023LI}.

In \citep{Edelman2022} and \citep{Trauger2023LI}, the authors derive covering number bounds for the matrix-valued linear function class $x\mapsto W^{T}x$ under various norm constraints. Their approaches rely on Maurey’s Sparsification Lemma. More specifically, under the corresponding norm constraints, their analyses yield the following covering number bound:
\begin{align}
\log
\mathcal{N}_{\infty}
\left(
\mathcal{F}^{(2,1)},
\epsilon,n,\|\cdot\|_2
\right)
\lesssim
\frac{B_w^2B_x^2}{\epsilon^2}
\log(kd).
\label{askmex}
\end{align}

By using a different approach that directly exploits the matrix structure, we obtain the following bound in Theorems \ref{thm:main0} and \ref{thm:main0b}:
\begin{align}
\log
\mathcal{N}_{\infty}
\left(
\mathcal{F}^{(2,1)},
\epsilon,n,\|\cdot\|_2
\right)
\lesssim
\frac{kd}{2}
\log\left(
\frac{16 B_x^2 B_w^2}{\epsilon^2}
\right).
\label{askmey}
\end{align}

Combining these two bounds, we obtain the following improved covering number estimate:
\begin{align}
\log
\mathcal{N}_{\infty}
\left(
\mathcal{F}^{(2,1)},
\epsilon,n,\|\cdot\|_2
\right)
\lesssim
\min\left\{
\frac{kd}{2}
\log\left(
\frac{16 B_x^2 B_w^2}{\epsilon^2}
\right),
\frac{B_w^2B_x^2}{\epsilon^2}
\log(k d)
\right\}.
\label{askme}
\end{align}
Furthermore, in Corollary 4.1.1 in \citep{Trauger2023LI}, the authors proposed the following bound for the same setting as Corollary \ref{cor:18}. 
\begin{align}
&\calR_n(\calF, \{X_{(i)}\}_{i=1}^n )\nn\\
& =O\bigg(B \bigg(\frac{B_x^3 \alpha}{\sqrt{n}}\bigg(1+ \log \bigg(\frac{\sqrt{n}}{B_x^2 \alpha}\bigg)\bigg) + B_x \sqrt{ \frac{\log 2d}{n}}\bigg)\bigg)
\end{align} where $B=B_w B_{w_c} L_{\sigma} B_{W_c}$ and $\alpha=B_{W_{QK}} \sqrt{2 \log(2d^2+1)}$. It is easy to see that this bound decays as $O(\frac{\log n}{\sqrt{n}})$. However, our bound in Corollary \ref{cor:18} decays as $O(\frac{1}{\sqrt{n}})$. Finally, our bound, like the result obtained by Trauger and Tewari, does not depend on the input length. 

Table \ref{tab:summary_comparison} provides a detailed comparison of our results with those of \citep{Trauger2023LI}, highlighting the differences in assumptions, dimension and sequence-length dependence, covering number bounds, and resulting generalization rates. 

\begin{table}[!t]
\centering
\caption{Summary of the main differences between our results and
existing covering number bounds.}
\label{tab:summary_comparison}
\begin{tabular}{lccc}
\hline
 & \textbf{Previous work}
 & \textbf{Ours: fixed subspace}
 & \textbf{Ours: rank constraint}
\\
\hline
Column-space constraint
& General
& $\operatorname{col}(W)\subseteq V_W$
& None
\\
Rank constraint
& --
& $\dim(V_W)=r_w$
& $\operatorname{rank}(W)\leq r_w$
\\
Explicit rank dependence
& No
& Yes
& Yes
\\
Dimension dependence
& $d,k$
& $d,r_w$
& $d+k-r_w, r_w$
\\
Dependence on input sequence length
& No
& No
& No
\\
Norm constraint
& Operator norm
& Operator norm
& Operator norm
\\
Covering bound
& $\displaystyle \frac{q_0^2}{\epsilon^2}$
& $\displaystyle
\min\left\{
a_1\log\left(\frac{b_1}{\epsilon^2}\right),
\frac{q_1^2}{\epsilon^2}
\right\}$
& $\displaystyle
\min\left\{
a_2\log\left(\frac{b_2}{\epsilon^2}\right),
\frac{q_2^2}{\epsilon^2}
\right\}$
\\
Generalization rate
& $\mathcal{O}(n^{-1/2} \log n)$
& $\mathcal{O}(n^{-1/2})$
& $\mathcal{O}(n^{-1/2})$
\\
\hline
\end{tabular}
\end{table}

\section{Conclusion}
In this paper, we developed a unified covering number framework for linear function classes under various norm constraints, with particular emphasis on column-space subspace constraints and rank constraints. Our results provide new covering number bounds that capture the geometric structure of the underlying matrix classes. In particular, when the prescribed subspace is the entire ambient space, our bounds yield novel estimates that, when combined with existing results, improve upon previously established covering number bounds under the same norm constraints. More generally, by explicitly incorporating rank constraints, our analysis quantifies how low-rank structures reduce the effective complexity of matrix classes and thereby affect their metric entropy.

\newpage
\appendices
\section{Proof of Theorem \ref{thm:main1}} \label{thm:main1:proof}
Let
\[
Q\in\mathbb{R}^{k \times r_w}
\]
be a matrix whose columns form an orthonormal basis of $V_W$. Thus,
\[
Q^TQ=I_{r_w},
\qquad
\operatorname{col}(Q)=V_W.
\]

Since
\[
\operatorname{col}(W)\subseteq V_W
\]
for every
\[
W\in\mathcal{W},
\]
there exists a unique matrix
\[
A\in\mathbb{R}^{r_w\times d}
\]
such that
\begin{align}
W=QA  \label{eq1}. 
\end{align}

Write
\[
W=[w_1,\ldots,w_d],
\qquad
A=[a_1,\ldots,a_d].
\]
Then, by \eqref{eq1}
\[
w_j=Qa_j,
\qquad j=1,\ldots,d.
\]
Since $Q^TQ=I_{r_w}$,
\[
\|Qa_j\|_2^2
=
a_j^TQ^TQa_j
=
\|a_j\|_2^2.
\]
Hence
\[
\|w_j\|_2=\|a_j\|_2,
\qquad j=1,\ldots,d.
\]
Therefore,
\begin{align}
\|W\|_{2,1}
&=
\sum_{j=1}^d\|w_j\|_2
\nonumber\\
&=
\sum_{j=1}^d \|a_j\|_2
\nonumber\\
&=
\|A\|_{2,1}.
\label{eq:l21-isometry}
\end{align}

Consequently, the correspondence $W=QA$ maps
$\mathcal{W}^{(2,1)}$ bijectively onto
\begin{align}
\mathcal{A}
=
\left\{
A\in\mathbb{R}^{r_w\times d}:
\|A\|_{2,1}\le B_w
\right\} \label{eq2}. 
\end{align}

Define the reduced function class
\[
\mathcal{G}
=
\left\{
g_A:x\mapsto Ax:
A\in\mathcal{A}
\right\}.
\tag{3}
\]

For every $W=QA$,
\[
f_W(x)
=
Wx
=
QAx
=
Qg_A(x).
\tag{4}
\]

Now let $A,A'\in\mathcal A$. For every $i=1,\ldots,N$,
\[
\begin{aligned}
\|f_{QA}(x_i)-f_{QA'}(x_i)\|_2
&=
\|Q(A-A')x_i\|_2.
\end{aligned}
\]
Since $Q^TQ=I_{r_w}$,
\[
\|Qz\|_2=\|z\|_2,
\qquad
z\in\mathbb{R}^{r_w}.
\]
Thus,
\[
\|f_{QA}(x_i)-f_{QA'}(x_i)\|_2
=
\|(A-A')x_i\|_2.
\]
Taking the maximum over $i$ gives
\begin{align}
\max_{1\leq i\leq n}
\|f_{QA}(x_i)-f_{QA'}(x_i)\|_2
&=
\max_{1\leq i\leq n}
\|(A-A')x_i\|_2.
\label{eq:empirical-isometry}
\end{align}

Therefore, the mapping
\[
g_A\longmapsto f_{QA}
\]
is an isometry with respect to the empirical uniform metric. Hence,
\begin{align}
\mathcal{N}_{\infty}
\left(
\mathcal{F}^{(2,1)},
\epsilon,n,\|\cdot\|_2
\right)
&=
\mathcal{N}_{\infty}
\left(
\mathcal{G},
\epsilon,n,\|\cdot\|_2
\right).
\label{eq:covering-isometry}
\end{align}

The reduced class $\mathcal G$ consists of linear maps
\[
g_A:\mathbb R^d \to\mathbb R^{r_w}
\]
whose parameter matrices satisfy
\[
\|A\|_{2,1}\le B_w.
\]
Moreover,
\[
\|x_i\|_1\le B_x,
\qquad i=1,\ldots,n.
\]
Thus, $\mathcal G$ is an $L_{2,1}$-bounded linear-function class
with matrix dimension $r_w\times k$.

Now, by applying Lemma \ref{lem:maurey}, Trauger et. al. \citep[Lemma 3.4]{Trauger2023LI} show that
\begin{align}
\log
\mathcal{N}_{\infty}
\left(
\mathcal{G},
\epsilon,n,\|\cdot\|_2
\right)
\lesssim
\frac{B_w^2B_x^2}{\epsilon^2}
\log(r_wd ),
\label{eq:trauger-reduced}
\end{align}
where the implicit constant is universal.

Combining \eqref{eq:covering-isometry} and
\eqref{eq:trauger-reduced}, we obtain
\[
\log
\mathcal{N}_{\infty}
\left(
\mathcal{F}^{(2,1)},
\epsilon,n,\|\cdot\|_2
\right)
\lesssim
\frac{B_w^2B_x^2}{\epsilon^2}
\log(r_w d).
\]
This completes the proof.

\section{Proof of Lemma~\ref{lem:small-epsilon-rank-entropy}}\label{lem:small-epsilon-rank-entropy:proof}
Let
\[
m=r_wd>1
\]
and define
\[
t=\frac{\epsilon^2}{B_x^2B_w^2}.
\]
Then $t\downarrow0$ as $\epsilon\downarrow0$, and
\[
\frac{16B_x^2B_w^2}{\epsilon^2}
=
\frac{16}{t}.
\]
Therefore, inequality \eqref{eq:small-epsilon-comparison} is
equivalent to
\[
\frac{m}{2}
\log\left(\frac{16}{t}\right)
<
\frac{1}{t}\log m,
\]
or, equivalently,
\[
\frac{mt}{2}
\log\left(\frac{16}{t}\right)
<
\log m.
\]

Since
\[
\lim_{t\downarrow0}
t\log\left(\frac{16}{t}\right)
=
0,
\]
we have
\[
\lim_{t\downarrow0}
\frac{mt}{2}
\log\left(\frac{16}{t}\right)
=
0.
\]
On the other hand,
\[
\log m>0
\]
because $m>1$. Hence, by the definition of a limit, there exists
$t_0>0$ such that
\[
\frac{mt}{2}
\log\left(\frac{16}{t}\right)
<
\log m
\]
for every $0<t\leq t_0$.

Choose
\[
\epsilon_0
=
\min\left\{
B_xB_w\sqrt{t_0},
\,2B_xB_w,
\,\frac{B_x}{2}
\right\}.
\]
Then
\[
0<\epsilon\leq\epsilon_0
\]
implies
\[
t
=
\frac{\epsilon^2}{B_x^2B_w^2}
\leq t_0.
\]
Consequently,
\[
\frac{mt}{2}
\log\left(\frac{16}{t}\right)
<
\log m.
\]
Substituting back
\[
m=r_wd,
\qquad
t=\frac{\epsilon^2}{B_x^2B_w^2},
\]
gives
\[
\frac{r_wd}{2}
\log\left(
\frac{16B_x^2B_w^2}{\epsilon^2}
\right)
<
\frac{B_w^2B_x^2}{\epsilon^2}
\log(r_wd).
\]
This proves the lemma.
\bibliographystyle{unsrt}
\bibliography{isitbib}
\end{document}